\documentclass[10pt,twocolumn,letterpaper]{article}
\usepackage{iccv}
\usepackage{times}
\usepackage{epsfig}
\usepackage{graphicx}
\usepackage{amsmath}
\usepackage{amssymb}
\usepackage{multirow}
\usepackage{subcaption}
\usepackage{bm}
\usepackage{soul,color}
\usepackage[linesnumbered,ruled,vlined,longend]{algorithm2e}
\usepackage[breaklinks=true,bookmarks=false]{hyperref}
\iccvfinalcopy 

\ificcvfinal\fi

\begin{document}

\title{Learning Driver Models for Automated Vehicles via Knowledge Sharing and Personalization}

\author{Wissam Kontar $^\mathsection$ \hspace{1cm} Xinzhi Zhong $^\mathsection$ \hspace{1cm} Soyoung Ahn $^{\mathsection,*}$\\
{\tt\small kontar@wisc.edu} \hspace{1cm} {\tt\small xzhong54@wisc.edu} \hspace{1cm} {\tt\small sue.ahn@wisc.edu}
\and 
$^\mathsection$Department of Civil and Environmental Engineering\\
University of Wisconsin-Madison, USA
}
\maketitle
\ificcvfinal\thispagestyle{empty}\fi

\begin{abstract}
   This paper describes a framework for learning Automated Vehicles (AVs) driver models via knowledge sharing between vehicles and personalization. The innate variability in the transportation system makes it exceptionally challenging to expose AVs to all possible driving scenarios during empirical experimentation or testing. Consequently, AVs could be blind to certain encounters that are deemed detrimental to their safe and efficient operation. It is then critical to share knowledge across AVs that increase exposure to driving scenarios occurring in the real world. This paper explores a method to collaboratively train a driver model by sharing knowledge and borrowing strength across vehicles while retaining a personalized model tailored to the vehicle's unique conditions and properties. Our model brings a federated learning approach to collaborate between multiple vehicles while circumventing the need to share raw data between them. We showcase our method's performance in experimental simulations. Such an approach to learning finds several applications across transportation engineering including intelligent transportation systems, traffic management, and vehicle-to-vehicle communication. \\
   Code and sample dataset are made available at the project page \href{https://github.com/wissamkontar}{https://github.com/wissamkontar}.
\end{abstract}

\section{Introduction}

The curse of variability stands as a critical barrier in the development and deployment of Automated Vehicles (AVs) in the open world. Variability in the open world stems from the innate dynamic, stochastic, and unpredictable nature of the transportation system. The driving task can change significantly depending on the traffic state (congestion, free flow, etc.), weather conditions (foggy, snowing, etc.), and roadway design (divided highway, one-way, etc.). It even depends on what other agents (pedestrians, bikers, busses, etc.) are present. Human drivers adapt and respond to these variations instinctively, but replicating such situational knowledge for an AV to maneuver safely in any given scenario, is extremely challenging in light of limited data availability and difficulty of real-world experimentation. Additionally, AVs are designed with some desired performance in mind. This means that AV's data can exhibit significant heterogeneity. This creates another challenge in learning driver models, as one needs to decouple the unique and common features from AV data to create personalized driver models for each vehicle. 

In a recent release of large-scale AV dataset from Waymo \cite{Kan_2023_arxiv} we see limited variability in the testing environment. For instance, the Waymo open dataset shows the majority of trip logs in sunny (99.3\%), daytime (80.6\%), and urban scenarios \cite{hu2023autonomous}. Another large dataset from nuScenes includes more diverse driving environments and scenarios as some driving logs come from different cities (Boston and Singapore) and locations (urban, residential, and industrial), and does include some rainy and cloudy weather conditions. However, most open world experiments and data collection (e.g., Cruise, Lyft, Aurora) are being done in dedicated routes and locations with limited exposure to variability in the transportation system. Dedicated experiments in similar driving scenarios allow for a deeper understanding of the AV performance in certain scenarios. However, a breadth of exposure is critically needed to train on extensive scenarios and encounters. 

The constraint on data availability is compounded by the difficulty of conducting real-world experiments. The development and testing of AVs require extensive resources and come with inherent safety concerns. Consequently, we see many experiments with exposure to limited scenarios, likely leaving AVs blind to a wide array of encounters. It is thus critical to develop methods to share knowledge across AVs to increase exposure to a wide range of scenarios occurring in the open world. 

Recent literature has also uncovered how AVs can exhibit different behaviors on the road based on the underlying design and control logic of these vehicles. For instance, we show in \cite{kontar2021multi,kontar2022bayesian}, how the car-following (CF) behavior of an AV exhibits a range of behavior depending on the underlying control parameters on spacing, desired speed, and acceleration constraints. Thus, data from AVs --specifically multi-class AVs (i.e., those designed with different performances in mind)-- is extremely heterogeneous. This heterogeneity is in fact a blessing rather than a curse since it gives us the opportunity to train a tailored model for each vehicle that considers its own unique characteristics. Such a process is referred to as personalization hereafter. 

This work is concerned with developing a framework through which different AVs can share knowledge from their encounters and borrow strength from each other, yet retain a personalized model tailored to their conditions and unique properties. A driver model here refers to a model capable of safely and effectively maneuvering the AV in various driving scenarios. This driver model controls the vehicle's perception, decision-making (on acceleration, braking, and steering), and navigation. However, the main scope of this work is in presenting how the sharing of information (through parameter transfer) while retaining a personalized model is done, and not on designing an optimal driver model.

\subsection{Motivation} \label{sec:motivation}
One can question the need to share knowledge through collaboration between vehicles as opposed to just pooling the data and learning one model. We argue that learning one-size-fits all can lead to misleading results, and dilutes the distinction between (i) different experienced driving scenarios, and (ii) heterogeneous driving behavior from subject vehicles. 

For instance, consider the basic unit of a driver model - the car following (CF) model - whereby an AV uses its sensor data to regulate its own acceleration/speed such that it follows its leader in  a safe manner. The design of such a CF controller for an AV requires specific consideration of desired spacing, speed preference, and comfortable acceleration. We can see such a model in play when we consider the widely adopted linear controller in Adaptive Cruise Control (ACC) and even full self driving systems \cite{shladover2015cooperative,milanes2014modeling,kontar2021multi}. In a linear controller, the desired spacing is first modeled based on the constant time gap policy as

\begin{equation}
\label{eq:tgp}
d_v^*(t) = v_v(t)\times\tau_v^* + \delta_v^*
\end{equation}
where $d_v^*(t)$ is the desired equilibrium spacing of vehicle $v$ at any time $t$; $v_v(t)$ is the respective speed of vehicle \textit{v}; $\tau_v^*$ is the constant time gap; and $\delta_v^*$ is the standstill distance. Accordingly, the deviation from the equilibrium spacing can be written as $\Delta d_v(t) = d_v(t) - d_v^*(t)$, where $d_v(t)$ represents the actual spacing between vehicle $v$ and its leader ($v-1$) at time $t$, and the speed difference between vehicle $v$ and its leader ($v-1$) is $\Delta v_v(t) = v_{v-1}(t) - v_v(t)$. 

As such, a system state for the AV controller can be described by \sloppy{$\bm x_v(t)=[\Delta d_v(t),\Delta v_v(t), a_v(t)]^T$} and input state as $u_v(t)$

\begin{equation}
\label{eq:acc}
  \begin{aligned}
    &u_v(t) = \bm K_v^T\bm x_v(t) \\        
    &\bm K_v^T = [k_{sv}, k_{vv}, k_{av}]
  \end{aligned}
\end{equation}
where $k_{sv}$, $k_{vv}$, $k_{av}$ are the feedback gains for the deviation from equilibrium spacing ($\Delta d_v(t)$), speed difference ($\Delta v_v(t)$) and acceleration ($a_v(t)$), respectively. The parameter setting of $\bm K_v^T$ denotes the regulation magnitude for each component in the system state $\bm x_v(t)$, and thus regulates the AV behavior. We refer readers to \cite{kontar2021multi} for an in-depth analysis on this phenomenon. 

It follows that one can design an AV with specific consideration of $d_v^*$, and $\bm K_v$ in mind. The dichotomy of these parameters is that they are (i) influenced by designer preference (or rider preference as shown in time headway parameter $\tau_v^*$ in commercial ACC), and (ii) influenced by the driving environment. The authors have a prior work that analyzes in depth how an AV can be designed with specific performance in mind, see \cite{kontar2021multi}. However, if data from heterogeneous AVs (i.e., AVs that differ in design and desired performance) is pooled together to train a single driver model, heterogeneity is lost since the training aims to achieve generalization and the underlying assumption is that data is homogeneous. One may work with stochastic models by estimating parameter distributions to regain some heterogeneity in the data. However, personalized models for each AV still cannot be tracked when the data are pooled. Accordingly, personalizing desired speed, headway, and spacing is unenforceable. 

It is important to note that in some cases data pooling is not even achievable. Commercial AV data can be protected by propriety rights and privacy concerns. Thus accessing raw data for training purposes may not be available. 

Accordingly, this work is motivated by the need to (i) share driving knowledge between different vehicles to increase the exposure of an AV to different driving scenarios/environments, (ii) retain a personalized model for each vehicle under heterogeneous behavior, and (iii) bypass the need to access raw data for training driving models. 

We follow the motivation discussed here with a simulation study (presented in Section \ref{sec:experiment2}) that shows how pooling data is not ideal under heterogeneity and how our approach can tackle this problem. 

\subsection{Related work}
With the advent increase of computational power and sheer amount of data collected in today's systems, federated learning became a powerful tool with the intent of \emph{processing the data where it was created on the edge}. The edge here refers to single device, vehicle, or the like. As a consequence, traditional Internet of Things (IoT) applications have shifted to a decentralized approach termed Internet of Federated Things (IoFT) \cite{kontar2021internet}. This brought about multiple analytical and computational tools that define how devices are set to collaborate with each other and share information. One of the early forefront tools in federated data analytics is the federated averaging (FedAvg), which was tailored for deep learning applications \cite{mcmahan2017communication}. The idea here is simple; devices in  a network structure would collaborate to learn a global deep learning model with the coordination of the central sever. Local devices perform iterations of stochastic gradient descent (SGD) using their data to obtain local parameters of their deep learning model, and send those parameters to a central server. Then the central server takes an average of those parameters to update the global model. Since then, several works have refined federated data analytics and tailored it to different applications. Notably, \cite{yue2021federated,yue2022federated} scales the application of federated learning into Gaussian processes, and general linear models. \cite{shi2021fed} presents Fed-ensemble, bringing ensemble methods to federated learning, improving generalization and uncertainty quantification. \cite{shi2022personalized} tailors a federated algorithm to learn unique and shared features for principal component analysis. And several other models exist, which are reviewed here \cite{kontar2021internet,wang2021field,bonawitz2022federated}. 

While the application of federated learning in transportation system has seen some momentum, it has yet to expand. Recently, \cite{chellapandi2023survey} presented a survey review of federated learning for connected and automated vehicles. They note several applications of federated learning for in-vehicle human monitoring, steering wheel prediction, object detection, motion control, and vehicle trajectory prediction. Most relevant to this work are the ones related to vehicle trajectory prediction. For example, \cite{han2022federated} uses an encrypted federated network algorithm to learn driver behavior and predict trajectories. \cite{rjoub2022explainable} designs a federated deep reinforcement learning for trajectory planning. However, only few efforts have been invested with limited scope, and the application domain of vehicle trajectory prediction via federated learning remains largely unexplored.

The focus of this paper is substantially different from available work and explores the potential of federated learning in a different direction. Specifically, what we aim to do is to share knowledge across vehicles to learn driving scenarios that might have been missed due to the variability in the transportation system, while also retaining personalization for each vehicle. 

\subsection{Main contribution}
We summarize our contributions in the following: 

\begin{itemize}
    \item \textbf{Modeling}: In our model, we acknowledge that data coming from each vehicle in uniquely heterogeneous given variability of driving scenarios, and personal preferences, and thus learning one-size-fits-all model is not ideal. Instead our model is personalized to encode unique encounter data and allows for transferring knowledge of unseen driving scenarios from one-vehicle to another. 
    \item \textbf{Algorithm}: We showcase a training algorithm based on federated learning, where vehicles only need to share iterations of personalized model parameters thus preserving privacy and minimizing communication cost.
    \item \textbf{Application}: We present an application of such framework on learning traffic oscillations by knowledge transfer between three vehicles, and another application on training personalized models under heterogeneity. 
\end{itemize}

The rest of the paper is organized as follows: Section \ref{sec:methodology} details our methodology. In Section \ref{sec:simulation} we introduce and analyze two simulation studies for knowledge sharing and personalization under heterogeneity in behavior. Finally, Section \ref{sec:conclusion} concludes. 

\section{Methodology}\label{sec:methodology}
In this section we discuss the problem setting and the model formulation. 

\subsection{Problem Setting}

\begin{figure}[!htb]
    \centering
    \includegraphics[width=0.8\linewidth]{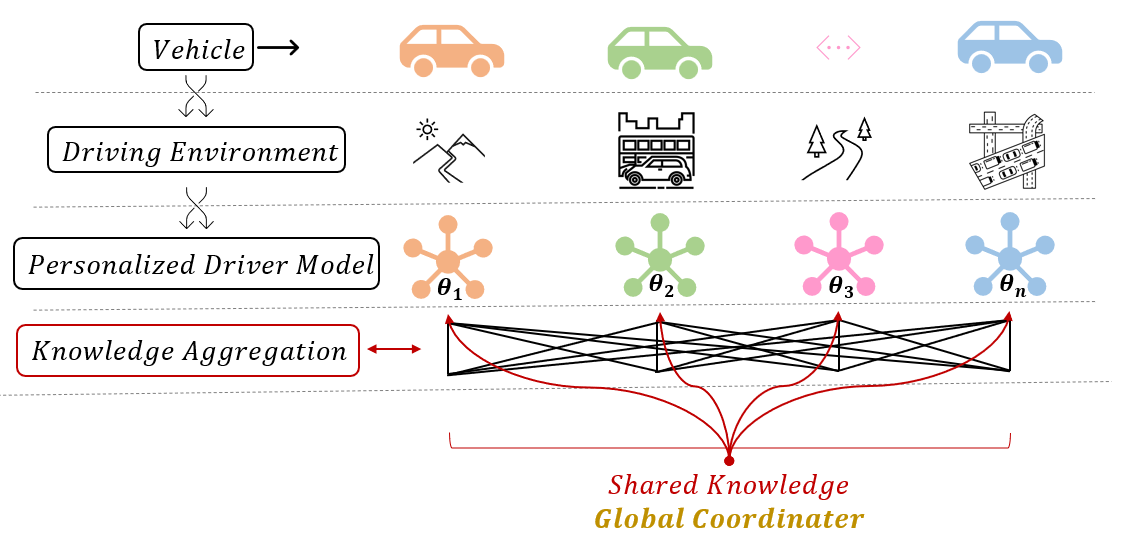}
    \caption{Problem Illustration}
    \label{fig:method}
\end{figure}

Consider the problem setting illustrated in Fig. \ref{fig:method}. Multiple vehicles are being tested in different driving environments, and each has a different personalized driver model. All of the driving encounters experienced by the vehicles are of interest to us, as the ultimate goal is to design a global driving model capable of maneuvering the vehicle in different driving environments. In perspective, to build such a model, one can extract the data for each vehicle alone, then train a local (i.e., using the vehicle's own data) driving model. This approach yields a single driving model for each vehicle that is blind to some driving encounters observed by other vehicles (due to different driving environments).

\begin{figure}[!htb]
    \centering
    \includegraphics[width=0.8\linewidth]{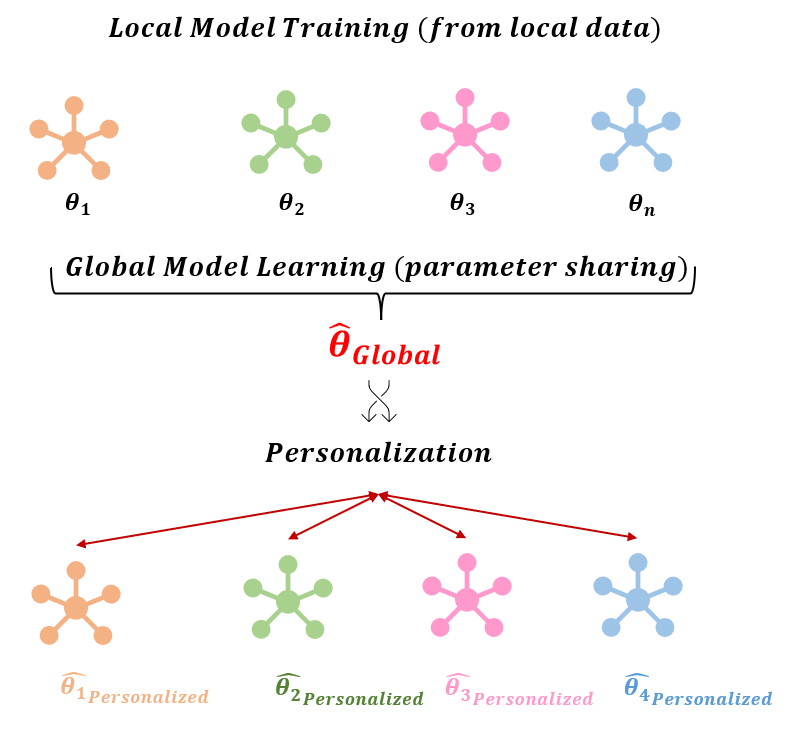}
    \caption{Learning Structure}
    \label{fig:architecture}
\end{figure}

What we seek to accomplish here is to allow vehicles to collaboratively train a global driving model, that allows for (i) a personalized model that focuses on their own local data, and (ii) knowledge sharing by transfer of information from one vehicle to another. The underlying assumption here is that data from each vehicle is uniquely heterogeneous. This heterogeneity is due to the unique encounter the vehicle is exposed to and its personalized design (e.g., specific desired speed, acceleration constraints, etc.). To achieve this we structure our driving model learning process as shown in Fig. \ref{fig:architecture}. 

\subsection{Problem Formulation}

Suppose there exist $v \geq 2$ vehicles. Each vehicle $v \in [V] := \{1,...,V\}$ has local dataset expressed as $D_v = \{\boldsymbol{X_v}, \boldsymbol{y_v}\}$ with the cardinality of $N_v$. We have an output $\bm y_v = [y_1, ..., y_N]^T$, and an input $\bm X_v = [x_1^T, ..., x_N^T]$. Here, $\bm X_v$ represents the state of the subject vehicle and nearby vehicles, usually a vector of data informed from on-board sensors. For instance, the state data can be relative speed between subject vehicle and leading vehicle, spacing, acceleration, jerk, and others. $\bm y_v$ represents an action by the AV, as acceleration magnitude for example. Formally, learning a driver model $f_v$ given training data $\bm X_v$ and $\bm y_v$ is a function defined as: 

\begin{equation}
   y_v(x_v) = f_v(x_v;\theta_v)
\end{equation}
where $f(*)$ can hold any functional form. Most notably, $f(*)$ can be a general deep learning network, Gaussian process, reinforcement learning, control model, or a linear model. The main interest here is parameter $\theta_v$, which parameterize $f(*)$ based on training data. The training data are ultimately tied to driving scenarios that vehicle $v$ was exposed to. One can notice that predicting an accurate output $f(x_v^*:\theta_v)$ ultimately depends on accurate estimation of $\theta_v$. Training $f(*)$ entails minimizing a general loss function, $\mathcal{L}(\theta_v;x_v;y_v)$. There exist several optimizers to solve the minimization problem of $\mathcal{L}$ (e.g., Adams \cite{kingma2014adam}, Stochastic Gradient Descent (SGD), etc.). However, most widely adopted is the SGD approach as it offers generalization power and is extremely efficient \cite{keskar2016large}. In typical fashion in SGD, the model training is performed in successive iterations. At each iteration of training $t$, a subset of training data (indexed by $\xi$) $\bm X_{v\xi}$ and $\bm y_{v\xi}$ is taken to update model parameters according to the below: 

\begin{equation}\label{eq:sgd}
    \theta_v^{(t+1)} \leftarrow \theta_v^{(t)} - \eta^{(t)}g_v(\theta_v^{(t)};\xi^{(t)})
\end{equation}
where $\eta^{(t)}$ is the learning rate and $g_v$ is the stochastic gradient of the loss function $\mathcal{L}$. The outcome is a driving model $f(x_v;\theta_v)$ parameterized by optimal $\theta_v$.

\subsection{Knowledge Sharing through Federated Learning}
The above is a general approach to learning a data-driven driver model, based on local data. However, in our approach we do not want to solely rely on vehicle $v$'s own data that is tied to the driving scenarios it was exposed to, but want to integrate knowledge from other AVs that might have encountered different driving scenarios. As such, we present a collaborative learning scheme that aims at learning a global parameter $\hat\theta$ that minimizes a global loss function defined as:  

\begin{equation}
    \mathcal{L}(\hat\theta) := \sum^{V}_{v=1}{\alpha_v{\mathcal{L}(\theta_{v};D_v)}}
\end{equation}

\noindent where $\alpha_v$ is the weight parameter for vehicle $v$ in the collaborative training scheme. $\alpha_v = \frac{N_v}{\sum_{v=1}^{V}{N_v}}$, such that $\sum_{v=1}^{V}{\alpha_v} = 1$. Consequently, at each communication round, each vehicle runs steps of SGD to estimate its local parameters: 

\begin{equation}\label{eq:sgdfl}
    \theta^{(t+1)}_v \leftarrow \theta^{(t)}_v - \eta^{(t)}g_v(\theta^{(t)}_v;\xi^{(t)}_v)
\end{equation}

Afterwards, the global coordinator aggregates the model parameters $\hat{\theta}$ according to the below rule and sends back $\hat{\theta}$ to each vehicle. Essentially, the global coordinator here averages out the local parameters coming out from each individual vehicle's driver model. 

\begin{equation} \label{eq:aggregation}
    \hat{\theta}  = \sum_{v=1}^{V}{\alpha_v\theta_v}
\end{equation}

In this scheme, we have all vehicles participating during the training rounds to send their local parameters to the global coordinator. Note that an underlying assumption is that all vehicles have the same functional form $f(*)$ of the driver model. Thereafter, the global coordinator sends back the updated parameters to each vehicle. Knowledge from each vehicle's driver model is shifted around to all collaborating vehicles. We summarize the above in Algorithms \ref{alg:1} \& \ref{alg:2}.

\begin{algorithm}[!h]
\SetAlgoLined
    \textbf{Data}: number of vehicles $V$, their initial model parameters $\theta$, number of sharing rounds $s\in S$, and weight parameters $\alpha_v$\\
\For{$s = 1:S$}
  {Global coordinator broadcasts $\theta$\;
  \For{$v \in V$}
  {$\theta^{(0)}_v = \theta$ \;
  Perform SGD (Algorithm \ref{alg:2}) to update local vehicle parameters\;}
  \emph{Aggregation:} $\hat{\theta_{s}} = \sum_{v=1}^{V}{\alpha_v\theta_v}$ \;
  Set $\theta = \hat{\theta_{s}}$\;}
  Return $\hat{\theta_S}$
\caption{Knowledge Sharing through Federation (Learning a Global Model)}
\label{alg:1}
\end{algorithm}

\begin{algorithm}[!h]
\SetAlgoLined
    \textbf{Data}: index $v$, learning rate $\eta$, number of local updates $U$, and vehicle model parameters $\theta_v^{(0)}$\\
\For{$t = 1:U $}
  {Perform random sampling to get mini-batch data $D_v$ and index it $\xi_{v}^{(t)}$\;
  $\theta^{(t+1)}_v \leftarrow \theta^{(t)}_v - \eta^{(t)}g_v(\theta^{(t)}_v;\xi^{(t)}_v)$
  }
  Return $\theta_v^{(U)}$
\caption{SGD for Local Vehicle Updates}
\label{alg:2}
\end{algorithm}

\subsection{Personalization}
Consequently, knowledge sharing was achieved by learning a global model parameterized by $\hat\theta$ and shared with all participating vehicles. It follows that each vehicle needs to personalize $\hat\theta$ based on its own local data. Through this process we encode heterogeneous behavior by each vehicle that is described by its own design parameters. Such personalization can be achieved in numerous ways. However the most commonly used is the regularization concept, where each vehicle uses its own local data $D_v$ to minimize a penalized least squares loss function defined as: 

\begin{equation}\label{eq:regularized}
\resizebox{1\linewidth}{!}{$min_{\theta_{personal}} \frac{1}{N_v}\sum_{n=1}^{N_v}\mathcal{L}(f(x_v;\theta_{personal}),y_v) - \omega ||\hat\theta - \theta_{personal}||_2^2$}
\end{equation}
where $\omega$ is a positive coefficient, and $\hat\theta$ is the global parameter learned from Algorithms \ref{alg:1} \& \ref{alg:2}. This approach personalized $\theta_{personal}$ to each vehicle $v \in V$ while retaining global knowledge by encouraging a solution close to $\hat\theta$. This regularization encourages $\theta_{personal}$ not to drift to far away from $\hat\theta$, and it shown statistically to reduce over fitting and the bias-variance trade off. Accordingly, each vehicle re-runs Algorithm \ref{alg:2}, with $g_v$ now based on the regularized loss function defined in Eq. \ref{eq:regularized}. Note that one does not necessarily need to do as many iteration steps ($S$) as done in Algorithms \ref{alg:1} \& \ref{alg:2} , but a fraction of the steps can suffice.

\section{Simulation Analysis} \label{sec:simulation}
We provide here a simulation experiment to demonstrate the applicability of our approach in (i) collaborative knowledge sharing to learn un-encountered driving scenarios, (ii) personalization of driving models under heterogeneity.

\subsection{Simulation Setup}
Consider that we have different AVs denoted by $v$. In here, we adopt a simple driver model setup as the goal is not to analyze the prediction performance of the underlying driver model, nor to design a driver model (several work in the literature exist on that), but rather to see how much knowledge sharing and personalization can be achieved through our proposed training structure. Accordingly, we define our driver model as a car-following model that predicts the AV speed (i.e., the output $y_v$) based on an input of leader speed ($x_v$). We use a Guassian Process Regression ($\mathcal{GP}$) to design such driver model. Note that our framework (knowledge sharing and personalization) works under any driver model, so users can replace the $\mathcal{GP}$ with a model of choice. We formulate the driver model as follows:

\begin{align}
    &f_v \sim \mathcal{GP}(m(\cdot),c(\cdot,\cdot))\\
    &y_v = f(x_v) + \epsilon, \epsilon^{i.i.d} \sim \mathcal{N}(0,\sigma_{\epsilon}^2)
\end{align}

\noindent where $x \in X $ is the input, and $m(\cdot): X \rightarrow \mathbb{R}$ is the prior mean function, $c(\cdot,\cdot): X \times X \rightarrow \mathbb{R}$ is the prior covariance function, and $\epsilon$ is the observational noise with variance $\sigma_{\epsilon}$. We further consider the zero mean function, and we assume the covariance function $c(\cdot,\cdot) =\sigma_{o}k(\cdot,\cdot)$ for a kernel function $k(\cdot,\cdot) : X \times X \rightarrow \mathbb{R}$, where $\sigma_{o}$ is the output variance. We adopt the well known Radial Basis Function (RBF) as our kernel that is parameterized by the length scale $l$. Accordingly, we denote $\theta_v = (\sigma_0, l, \sigma_{\epsilon})^T \in \mathbb{R}^3$ the parameters to be estimated. To estimate $\theta_v$, we use SGD (Algorithm \ref{alg:2}) to minimize the scaled negative log marginal likelihood function defined as

\begin{equation}
    \mathcal{L}(\theta_v, X_v, y_v) = \frac{-1}{n}logp(y_v|X_v,\theta_v)
\end{equation}

Accordingly, the collaboration here is based on $\theta_v = (\sigma_0, l, \sigma_{\epsilon})^T \in \mathbb{R}^3$. At every sharing round $S$ (see Algorithm \ref{alg:1}), each vehicle performs steps of SGD on its $\mathcal{L}(\theta_v, X_v, y_v)$ to output a local set of parameters $\theta_v$ (see Algorithm \ref{alg:2}). At the aggregation step, a global coordinator then computes $\hat\theta$ based on the law described in Eq. \ref{eq:aggregation}, and sets $\theta_v = \hat\theta$. This scheme of collaborative learning and sharing of parameters is what encodes knowledge transfer between participating vehicles. It is important to note that at no point during the collaboration, raw data ($X_v$ and $y_v$) are shared between vehicles, which makes this approach very computationally efficient and privacy aware. Additionally, the parameters  $\theta_v$ are dependent on the assumed functional form $f(*)$ of the driver model. For instance, if one is using neural network predictor, then $\theta_v$ would be the weights of the neural net. 

Finally, each vehicle re-runs Algorithm \ref{alg:2} based on the regularized loss function described in Eq. \ref{eq:regularized} to compute the final $\theta_{personalized}$ used in the prediction of AV speed (i.e., predictions from the driver model).

\subsection{Experiment 1: Knowledge Sharing to Learn Traffic Oscillations}\label{sec:experiment1}
\begin{figure}[!htb]
     \centering
     \begin{subfigure}[b]{0.35\textwidth}
         \centering
         \includegraphics[width=\textwidth]{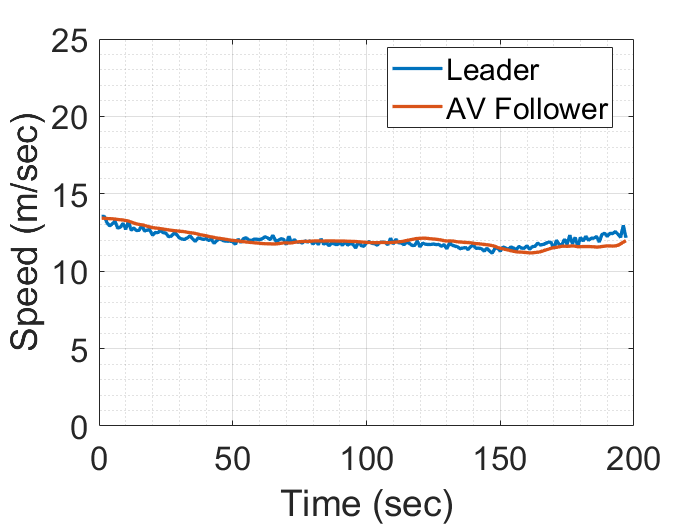}
                  \caption{Vehicle 1: Constant speed scenario}
     \end{subfigure}
     \hfill
     \begin{subfigure}[b]{0.35\textwidth}
         \centering
         \includegraphics[width=\textwidth]{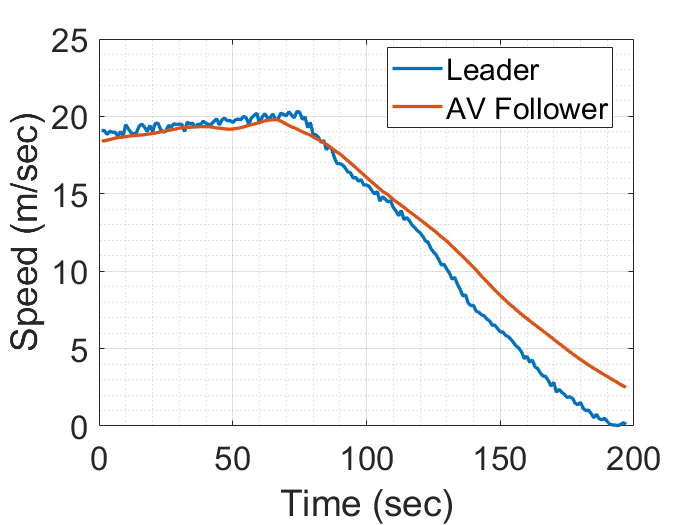}
                  \caption{Vehicle 2: Deceleration scenario}
     \end{subfigure}
     \hfill
     \begin{subfigure}[b]{0.35\textwidth}
         \centering
         \includegraphics[width=\textwidth]{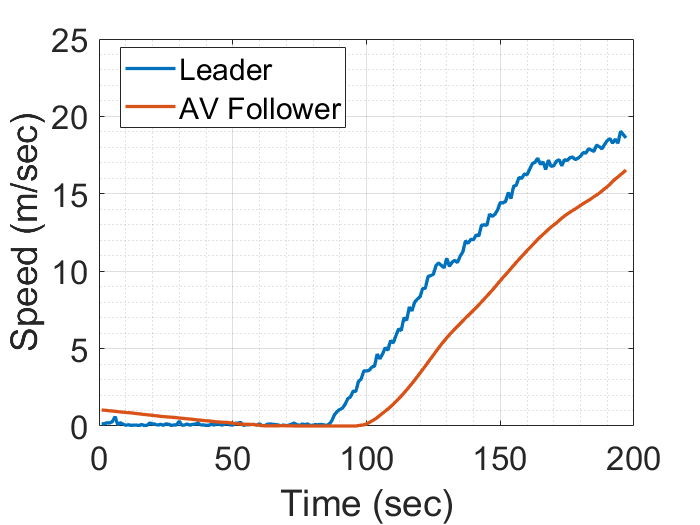}
                  \caption{Vehicle 3: Acceleration scenario}
     \end{subfigure}

\caption{Driving Scenarios Observed by each Vehicle}
\label{fig:setup}
\end{figure}

Now we consider a specific experimental setup for which we want to demonstrate how knowledge is transferred between vehicles. Consider now that we have three automated vehicles ($V=3$). We assume each vehicle was operating under completely different driving environment. To signify this, we consider three different scenario-based datasets shown in Fig. \ref{fig:setup}: (i) vehicle 1 is operating under constant speed, (ii) vehicle 2 experiences a deceleration maneuver, and (iii) vehicle 3 experiences an acceleration maneuver. Note that these experiments are extracted from Waymo dataset \cite{hu2022processing}. We use this dataset to give realism to our experimental setup. Each of the three driving scenarios runs for $19.7$-seconds long run at a $10Hz$ resolution. Consequently, each vehicle has a local dataset of $N_v = 197$. In Fig. \ref{fig:setup} the top row shows the driving scenario for each vehicle. The blue curve shows a human driver (leader) oscillation, and the red curve shows the resulting response of AV follower (the Waymo vehicle). Recall that our driver model is based on speed input and output. Thus, training our model requires an input, $X_v$, of the leading vehicle (blue curve) speed. And predicts an output speed $y_v$ of an AV (the red curve).

\begin{figure}[!htb]
\centering
\includegraphics[width=0.8\linewidth]{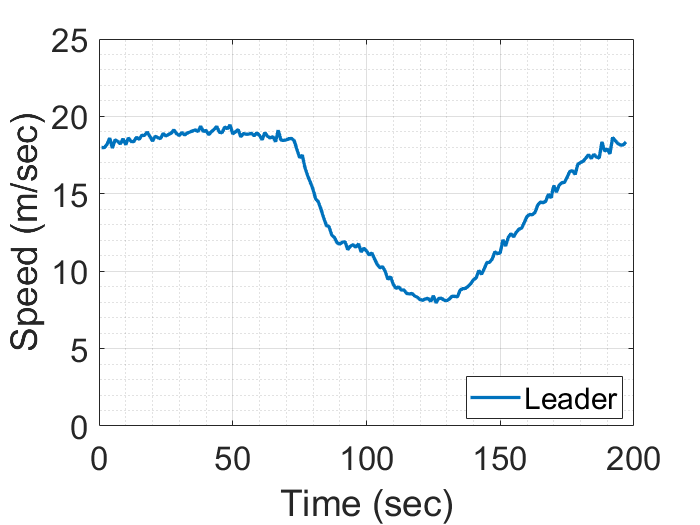}%
\caption{Testing Scenario: Full Oscillation}
\label{fig:osc}
\end{figure}

The three driving encounters in our setup (Fig. \ref{fig:setup}), represent portions of a full traffic oscillation -- constant speed followed by a deceleration and acceleration maneuver to reach a constant speed again. Such oscillations are ubiquitous in traffic systems and are certain to occur in the open world. The goal here is to allow the three vehicles to train a driving model in a collaborative fashion, in such a way that they would individually be able to respond to a traffic oscillation knowing that none of them had seen such a driving scenario. After collaborative training, we test each vehicle on the full oscillation scenario, seen in Fig. \ref{fig:osc}. The full oscillation in Fig. \ref{fig:osc}, represents an observed empirical oscillation created by a human driven vehicle (HDV) -- extracted from the Waymo dataset. 

\subsubsection{Simulation Results}

\begin{figure}[!htbp]
    \centering
    \begin{subfigure}{0.35\textwidth}
        \includegraphics[width=\textwidth]{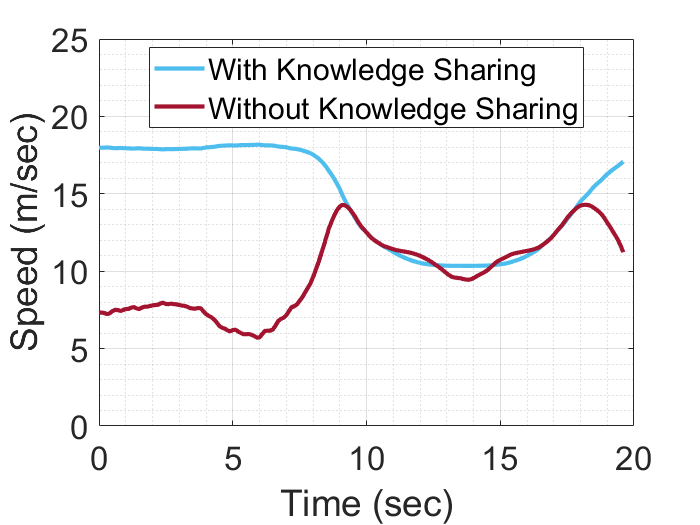}
        \caption{Vehicle 1}
        \label{fig:sub1}
    \end{subfigure}
    \hfill
    \begin{subfigure}{0.35\textwidth}
        \includegraphics[width=\textwidth]{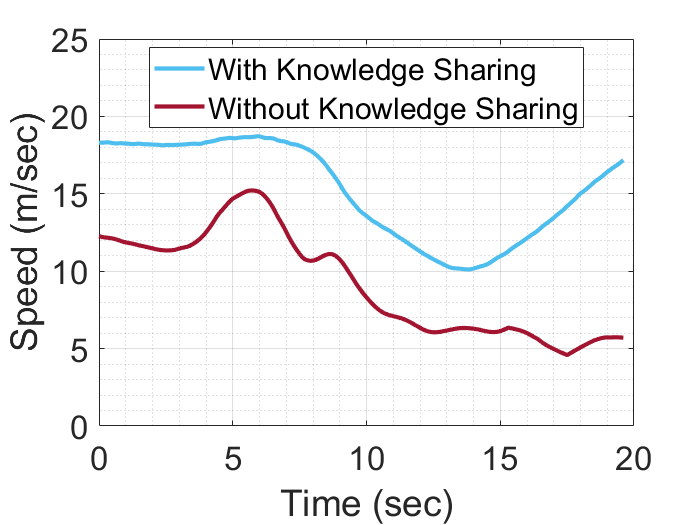}
        \caption{Vehicle 2}
        \label{fig:sub2}
    \end{subfigure}
    \hfill
    \begin{subfigure}{0.35\textwidth}
        \includegraphics[width=\textwidth]{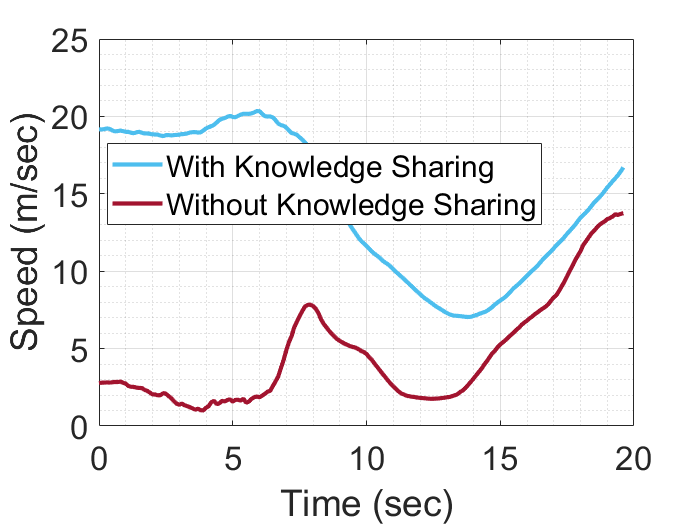}
        \caption{Vehicle 3}
        \label{fig:sub3}
    \end{subfigure}
    \caption{Prediction Results of each Vehicle on the Testing Scenario (Full Oscillation): Before and After Knowledge Sharing}
    \label{fig:knowledgesharing}
\end{figure}

After training the driver model with our approach that focuses on knowledge sharing and personalization, we test against the oscillation shown in Fig. \ref{fig:osc}, whereby the input is the leader speed, and output is the AV speed. The goal here is to see whether each vehicle (1-2-3) is able to produce an oscillation. Results are shown in Fig. \ref{fig:knowledgesharing}. Interestingly, we see that ``After Knowledge Sharing" (blue curve) each of the vehicle was able to produce a traffic oscillation even though none of the vehicles have full knowledge of an oscillation (recall Fig. \ref{fig:setup}). This is not the case when we look at ``Without Knowledge Sharing" (red curve). However, it is notable that the oscillations ``After Knowledge Sharing" are not perfect. This is rather expected given the limited training data, as the focus is not on building a driver model rather on knowledge sharing. In practice, usually a driver model would have a complex set of input based on speed of leader, position of leader, and multiple other factors. But, success here is denoted as the ability of each vehicle (1-2-3) to produce an AV response of an oscillation when subjected to one. 

The effect of knowledge sharing is specifically amplified when looking at vehicle 1 profile in Fig. \ref{fig:knowledgesharing}. Remember that vehicle 1 had only access to its local data which is a constant speed profile. Specifically, it shows that the driver model of vehicle 1 can extract knowledge on deceleration and acceleration phases from vehicles 2 \& 3, respectively. This gives it knowledge to respond to an oscillation. On the contrary, when we train the driver models of vehicles 1-2-3 without knowledge sharing, we notice that they can be blind to some maneuvers (deceleration/acceleration) that inhibits their ability to fully produce an oscillation, see red curves in Fig. \ref{fig:knowledgesharing}.

\subsection{Experiment 2: Knowledge Sharing and Personalization under Heterogeneity} \label{sec:experiment2}
Now we revisit the problem of data pooling and heterogeneity in AV behavior from the \emph{Motivation} (Section \ref{sec:motivation}). We consider here two AVs, each having a different design setup and thus driving behavior. Note that such design of AVs is not uncommon in the real world, see \cite{kontar2021multi}. 

\begin{figure}[!htb]
       \centering
        \includegraphics[width=0.8\linewidth]{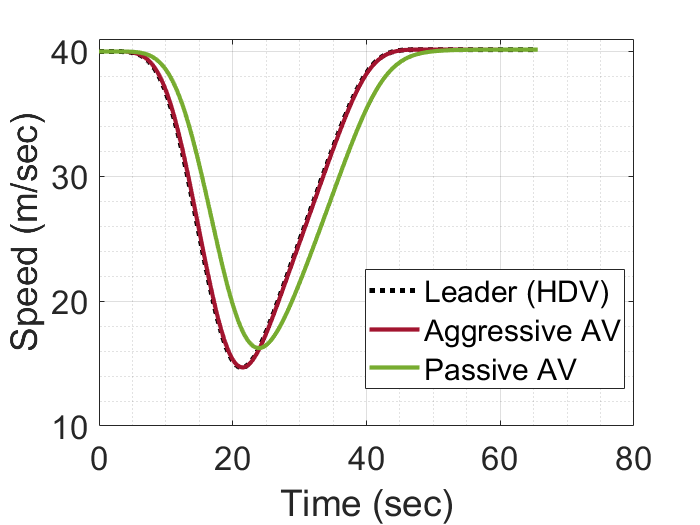}%
\caption{Speed Profiles for Experiment 2 Setup. Note: Leader HDV (black) and Aggressive AV (red) Curves Nearly Overlap each other}
\label{fig:heterogenous}
\end{figure}

\begin{itemize}
    \item Aggressive AV: is an AV that is designed to be very responsive to the leader speed and prioritizes the minimization of speed different $\Delta v_v(t)$. Specifically, for the aggressive AV we set $\bm K_v = [0.01, 10, -0.01]$, with $\tau^*_v = 0.5$, and $\delta_v^* = 5m$. 
    \item Passive AV: is an AV that is designed to very passive to the leader speed and prioritizes the minimization in the deviation from target spacing $d_v^*(t)$. Specifically, we set $\bm K_v = [10, 0.01, -0.01]$, with $\tau^*_v = 2.5$, and $\delta_v^* = 7m$. 
\end{itemize}

Further, we consider a specific leader (HDV) speed oscillation and then based on the linear controller explained in Section \ref{sec:motivation} and for the settings described above, we simulate an AV speed profile. This is shown in Fig. \ref{fig:heterogenous}. Note that the leader (HDV) profile is common between the two generated AV profiles. One can directly notice the difference in behavior between the two AVs. The aggressive AV (red) nearly masks its leader (black), while the passive AV (green) is much less reactive. It follows then that the data (i.e., output $y_v$) generated from control design of these two AVs exhibit heterogeneous features. 

\subsubsection{Simulation Results}

\begin{figure}[!htb]
       \centering
        \includegraphics[width=0.8\linewidth]{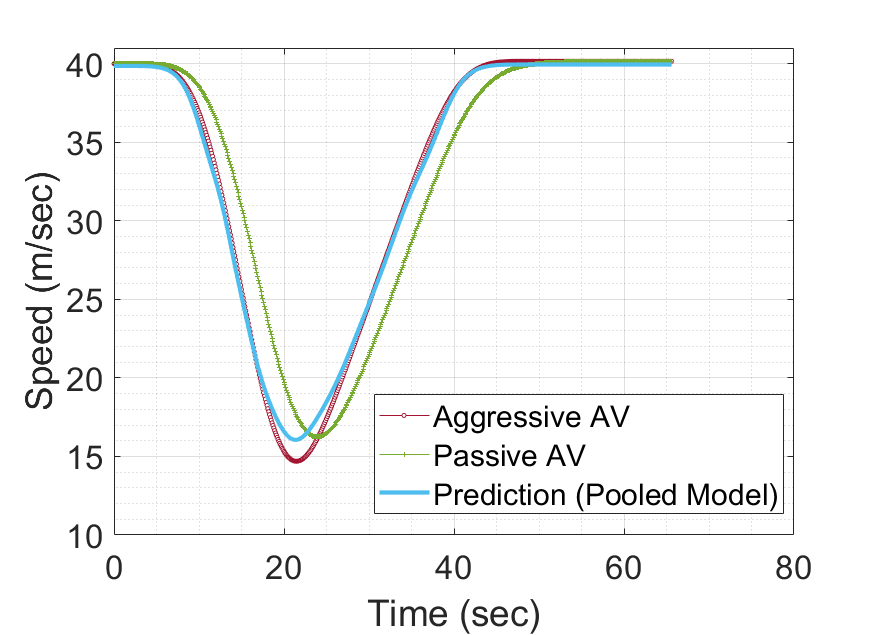}%
\caption{Prediction from a Driver Model based on Pooled Data}
\label{fig:pooled}
\end{figure}

When output data from these two heterogeneous vehicles is pooled together to learn one global driving model, it masks the underlying difference in behavior. This is further illustrated in Fig. \ref{fig:pooled}. When we train a driver model based on pooling the speed data from the aggressive and passive AVs, we get a behavior shown in Fig. \ref{fig:pooled} (light blue color). The prediction fails to distinguish between an aggressive and passive behavior as it tries to achieve generalization. 

\begin{figure}[!htb]
     \centering
     \begin{subfigure}[b]{0.35\textwidth}
         \centering
         \includegraphics[width=\textwidth]{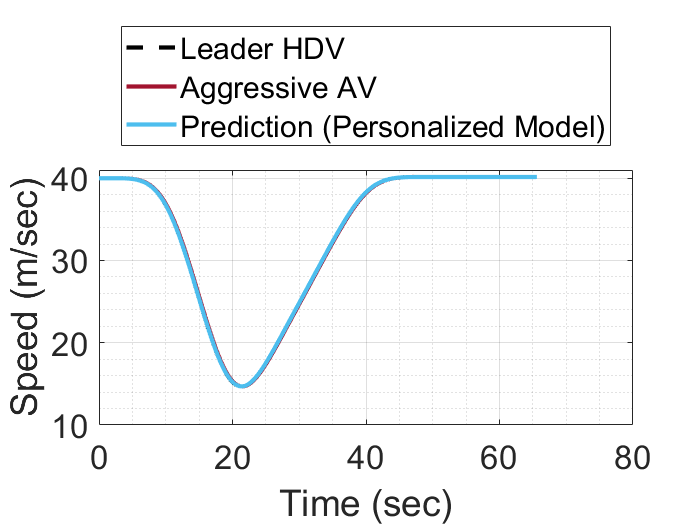}
         \caption{Personalized Aggressive Driver Model. Note: the Three Curves Nearly Overlap each other}
         \label{fig:personal_aggressive}
     \end{subfigure}
     \hfill
     \begin{subfigure}[b]{0.35\textwidth}
         \centering
         \includegraphics[width=\textwidth]{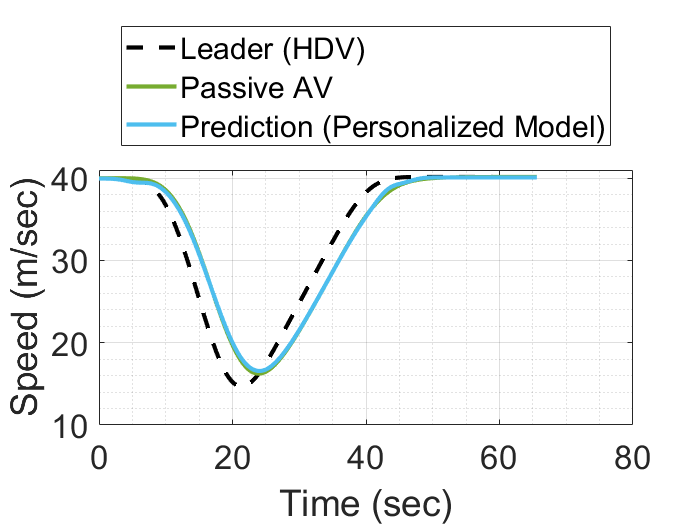}
         \caption{Personalized Passive Driver Model. Note: Blue and Green Curves Nearly Overlap each other}
         \label{fig:personal_passive}
     \end{subfigure}

\caption{Prediction from Driver Models based on Knowledge Sharing and Personalization}
\label{fig:personalized}
\end{figure}

However, in our proposed learning structure, the personalization step produces $\theta_{personalized}$ that results in different driver models for each vehicle rather than one-model-fits-all approach. This allows vehicles to retain their desired behavior which still sharing knowledge. Fig. \ref{fig:personalized} shows the predictions for both aggressive and passive AVs based on our personalized knowledge sharing approach. Table \ref{tab:error} further shows the Root Mean Squared Error (RMSE) based on speed for each model. It is evident that under heterogeneity our proposed learning structure can better encode different driving behaviors and thus better fit personalized driver models. 

\begin{table}[!htb]
    \centering
    \begin{tabular}{l l l}
        RMSE(speed) & Aggressive & Passive  \\ \hline
        Pooled Model & 0.53 & 2.46 \\ \hline
        Knowledge Sharing \& Personalization & 0.12 & 0.22 \\ \hline
    \end{tabular}
    \caption{Prediction Error}
    \label{tab:error}
\end{table}

\subsection{Limitations and Remarks}

\begin{enumerate}
    \item \textbf{On the driver model function:} Here we explored a $GP$ function; however in practice other models might be more widespread such as deep learning or reinforcement learning methods. While our training framework remains malleable enough to apply for different functions, it is important to extend to other driver models. 
    \item \textbf{On parameter sharing:} In our framework, all vehicles share the same driver model ($GP$) with same feature input (leader speed) and output (AV speed). However, if one wants to collaboratively share knowledge between vehicles with different functional form of the driver model (i.e., one vehicle has neural network and other reinforcement learning models) an alternative parameter sharing scheme is needed. A way to approach this is to decompose $\theta_s := \theta_{functional} + \theta_{shared}$. $\theta_{functional}$ are unique parameters that relevant to the vehicle driver model, while $\theta_{shared}$ come from a global learned model. Additionally, one can regularize the $\theta$ (i.e., assign weights on $\theta_{functional}$ and $\theta_{shared}$) so that a vehicle can put more value on its own driver model and preferences.
    \item \textbf{On aggregation ($\hat\theta$ in Eq. \ref{eq:aggregation}):} Our aggregation strategy (averaging) is based on the widely used FedAvg. However, other strategies exist and can be designed to suit the desired application. 
    \item \textbf{On personalization under complex driver models:} We note that in our approach here we use a very simplified driver model (only taking speed data), however in complex driver models it can become hard to encode personalization for each vehicle. Since in this case, you can have multiple layers of designed parameters that are all contributing to the behavioral change of the vehicle. This is rather a complex problem to solve, as one would first need to decompose unique and common features between vehicles. This problem is currently understudy by the team. 
    \item \textbf{On when to pool data:} In some cases pooling data can be beneficial. For instance, consider the setup in Experiment 1 (Section \ref{sec:experiment1}). If vehicles 1-2-3 have similar design (driver model with same parameter setting), one can pool all the data from different scenarios and train one-modal-that-fits-all. However, pooling data might not even be applicable given that access to some data is restricted (privacy concerns of propriety rights). As such our approach circumvents this by only sharing parameter values and never raw data.  
\end{enumerate}

\section{Conclusions} \label{sec:conclusion}
 In this work we present a way of learning and training driver models for AVs in a collaborative way. Different vehicles can share knowledge between each other through a collaborative iterative process that entails sharing and discovering optimal parameters that minimize a desired global loss function. We also show how vehicles can share knowledge while retaining a personalized model tailored to their own data. 

We showcase two experimental applications of the designed model. In the first experiment, three vehicles collaborate to learn a speed oscillation, by decomposing and transferring knowledge between each other. In the second experiment, we show how under heterogeneous AV behavior, learning a driver model while pooling data is not ideal and thus personalization yields better results.

Several limitations and extensions of this work are yet to be tackled and are highlighted above. A large scale experimentation and benchmarking of driver models with or without knowledge sharing and personalization is important, yet goes beyond what this paper can provide, and is left for future work by the authors. We hope that our endeavors in this modeling direction would spur interest and motivate further work on designing complex AV driver model's that can safely and efficiently maneuver in the open world.

\section{Acknowledgements}
This research was sponsored by the Unites States National Science Foundation through Award CMMI 1932932 and the University of Wisconsin-Madison. 

{\small
\bibliographystyle{ieee_fullname}
\bibliography{references}
}

\end{document}